\documentclass{article}

\PassOptionsToPackage{numbers, compress}{natbib}

\usepackage[final]{nips_2018}

\usepackage[utf8]{inputenc} 
\usepackage[T1]{fontenc}    
\usepackage{hyperref}       
\usepackage{url}            
\usepackage{booktabs}       
\usepackage{amsfonts}       
\usepackage{nicefrac}       
\usepackage{microtype}      

\usepackage{lmodern}

\usepackage[british]{babel}
\usepackage{csquotes}

\usepackage{listings, multicol}
\usepackage{color}
\definecolor{dkgreen}{rgb}{0,0.6,0}
\definecolor{gray}{rgb}{0.5,0.5,0.5}
\definecolor{mauve}{rgb}{0.58,0,0.82}
\usepackage{graphicx}
\usepackage{caption}
\usepackage{subcaption}
\usepackage{float}

\usepackage{amsmath}
\usepackage{amssymb}
\usepackage{proba}

\newcommand{\norm}[1]{\left\lVert#1\right\rVert}

\usepackage{bm}
\let\vec\mathbf

\usepackage{mathtools}

\usepackage{breqn}

\usepackage{verbatim}

\usepackage{siunitx}

\usepackage{enumitem}

\newcommand*{\thead}[1]{\multicolumn{1}{c}{\bfseries #1}}

\usepackage{booktabs}
\usepackage{multirow}

\usepackage{placeins}

\usepackage{chngpage}

\usepackage{pdfpages}

\newcommand{\DeepHitPlus}{\-DeepHit\textsuperscript{+}}

\title{Feature Selection for Survival Analysis with Competing Risks using Deep Learning}

\author{Carl Rietschel$^\dagger$, Jinsung Yoon$^*$ and Mihaela van der Schaar$^{\dagger*}$ \\
$^\dagger$University of Oxford, UK. $^*$University of California Los Angeles, USA\\
\texttt{carlrietschel@gmail.com}, \texttt{jsyoon0823@gmail.com}, \texttt{mihaela@ee.ucla.edu}\\
}

\begin{document}

\maketitle

\begin{abstract}
Deep learning models for survival analysis have gained significant attention in the literature, but they suffer from severe performance deficits when the dataset contains many irrelevant features. We give empirical evidence for this problem in real-world medical settings using the state-of-the-art model DeepHit. Furthermore, we develop methods to improve the deep learning model through novel approaches to feature selection in survival analysis. We propose filter methods for \textit{hard} feature selection and a neural network architecture that weights features for \textit{soft} feature selection. Our experiments on two real-world medical datasets demonstrate that substantial performance improvements against the original models are achievable.
\end{abstract}

\section{Introduction}
Recent research has produced a variety of successful new deep learning models for survival analysis. Whilst some methods \citep{Katzman2016, Zhu2016, Luck:2017} have strong parametric assumptions, more general models \citep{nemchenko2018siamese, giunchiglia2018rnn, Lee2018DeepHitAD} have been developed. However, deep learning approaches suffer from performance deficits when there are many irrelevant features. This can certainly be the case in medical datasets, where numerous features may be recorded about a patient (e.g. the PLCO dataset we use in this work - see details in Appendix \ref{app:datadescription}). In this paper, we give evidence for this problem using DeepHit\footnote{We thank the authors of DeepHit for sharing their source-code implementation with us} \citep{Lee2018DeepHitAD} on large real-world medical datasets, and propose feature selection techniques to achieve substantial performance improvements.

\section{Survival analysis with \DeepHitPlus{}}
We consider survival analysis with $K$ competing risks, focusing on the medical domain with right-censored data, as is frequently encountered in time-limited medical trials. Survival data for each of $N$ patients is a tuple $(\vec{x}, \tau, \delta)$: \textit{Covariates} $\vec{x}$ are the characteristics of each patient observed. \textit{Time} $\tau$ is measured from when the covariates were collected until the first event or censoring. \textit{Label} $\delta \in \{\emptyset, 1, \dots, K\}$ indicates which event (or right censoring, denoted $\emptyset$) occurred at time $\tau$. For each event $k$ the cause-specific \textit{cumulative incidence function} (CIF) gives the probability that this event occurs on or before time $t$ for a patient with covariates $\vec{x}$, and is denoted $F_k(t \mid \vec{x}) = \probX{\tau \leq t, \delta = k \mid \vec{x}}$. A machine learning model will attempt to give empirical estimates $\hat F_k (t \mid \vec{x})$ of the true cumulative incidence functions $F_k$.

In this paper we use DeepHit because it is very general, allows for competing risks and shows good empirical performance \citep{Lee2018DeepHitAD}. We furthermore develop \DeepHitPlus{} to implement two improvements: We switch the early stopping criterion from validation loss to the performance measure C-index, and allow for random search on the sizes (number of layers and hidden dimension) of the shared and cause-specific sub-networks.

\section{Automatic feature selection for \DeepHitPlus{}}
We propose models with \textit{filter} approaches to \textit{hard} feature selection, as well as reweighting features in \textit{soft} feature selection techniques, using \DeepHitPlus{}'s architecture as a basis for our methods. We also develop a \textit{hybrid} feature selection approach in Appendix \ref{app:hybriddeephitplus}, but do not investigate wrapper feature selection techniques due to their computational complexity \citep{kordos2017data}.

\subsection{Filter\DeepHitPlus{}: Automatic feature selection using filter method}
\begin{figure}
\centering
\begin{minipage}{.5\textwidth}
  \centering
  \includegraphics[width=\textwidth, page=1]{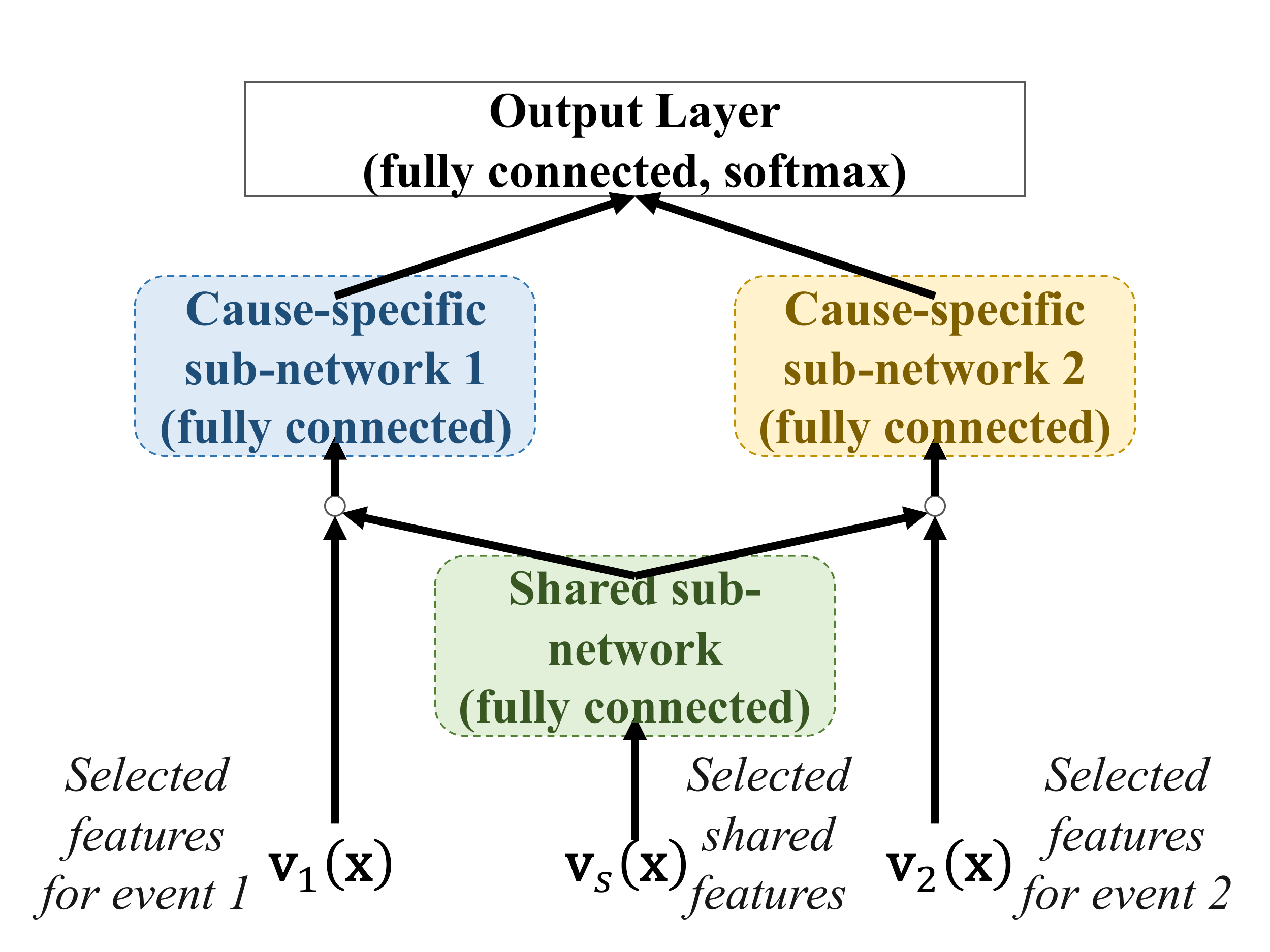}
  \captionof{figure}{Filter\DeepHitPlus{} architecture}
  \label{fig:dh_architecture_modified}
\end{minipage}%
\begin{minipage}{.5\textwidth}
  \centering
  \includegraphics[width=\textwidth, page=2]{figures/dh_architecture_nips.pdf}
  \captionof{figure}{Sparse\DeepHitPlus{} sub-network}
  \label{fig:dh_architecture_sparse}
\end{minipage}
\end{figure}

Whilst it would be possible to feed a subset of features in place of $\vec{x}$ into \DeepHitPlus{}'s original network, we desire the ability to make different feature selections for each competing risk. Filter\DeepHitPlus{} thus uses the modified architecture shown in Figure \ref{fig:dh_architecture_modified}, where the input connections to the shared and cause-specific sub-networks are replaced with certain selected subsets of the features. We denote these $\vec{v}_s(\vec{x})$ for the shared sub-network and $\vec{v}_k(\vec{x}), k=1,2,...$ for the $k$th cause-specific sub-network.

We cannot directly apply classical filter feature selection methods for classification or regression tasks to survival analysis: due to the right-censored data, the label is not determined for some patients. However, when fixing the time horizon to $\Delta t$, survival analysis becomes a classification task for which we can use filter techniques. In particular, if for patient $i$ time $\tau^{(i)}$ and label $\delta^{(i)}$ are observed, the binary cause-specific time-fixed label would be
\begin{equation}
\delta^{(i, \Delta t, k)} =
\begin{cases}
1, & \text{if } \tau^{(i)} < \Delta t \text{ and } \delta^{(i)} = k \\
0, & \text{otherwise}. \\
\end{cases}
\end{equation}
For each event, we apply the filter method multiple times to score associations between features and the cause-specific label for a preselected set of evaluation time horizons. The final score for each feature is then the average of its scores across the time horizons, and is used for automatic feature selection for the survival model. Our method in Filter\DeepHitPlus{} automatically selects the features for each cause-specific sub-network $\vec{v}_k(\vec{x})$ from the feature ranking. It chooses the top $m_k$ features according to the ranking, treating the number of features to be selected ($m_k, k=1,\dots,K$) as an additional hyperparameter optimized through random search. The shared sub-network's input features $\vec{v}_s$ are the common features that occur in all $\vec{v}_k$ (i.e. their intersection). If there are none, Filter\DeepHitPlus{} does not have a shared sub-network.

We desire methods that are able to deal with both continuous and categorical features, and choose three common filters for three new models: Analysis of variance / $t$-tests (\textbf{ANOVA}) \cite{Saeys:2007:FSInterpretability}, weights of a trained support vector machine (\textbf{SVM}) \citep{Bron:2015:SVMfeatureselection} and the \textbf{ReliefF} algorithm \citep{Kononenko1997}.

\subsection{Sparse\DeepHitPlus{}: Automatic feature selection using sparse initial layer} \label{sparsedeephitplus}
Inspired by linear models that achieve low or zero weights through introducing $L_1$ regularization, \cite{Li:2016aa} proposes adding an additional sparse layer with one-to-one connections from the input before the first hidden layer of a deep neural network. Our architecture for Sparse\DeepHitPlus{} applies this to the multitask learning setting of \DeepHitPlus{}, incorporating additional sparse layers for the shared sub-network as well as each of the cause-specific sub-networks. In Figure \ref{fig:dh_architecture_sparse}, we show this by illustrating the nodes, and connections between nodes in each layer.

Mathematically, we denote the input to the shared network $\vec{x}$, and the inputs to the $k$th cause-specific network $\vec{z}_k$. Then we define parameters $\vec{w}_s$ and $\vec{w}_k, k=1,\dots,K$, and the activations of the sparse layers are given by
\begin{align}
\vec{z}_s^' &= \vec{x} \odot \vec{w}_s \\
\vec{z}_k^' &= \vec{z}_k \odot \vec{w}_k, \quad k=1,\dots,K,
\end{align}
where $\odot$ denotes the element-wise (\textit{Hadamard}) product of two vectors. These outputs $\vec{z}_s^'$ and $\vec{z}_k^'$ are then the inputs to the fully connected parts of the sub-networks as before.

The loss function, originally $\mathcal{L}$, is adjusted to include additional $L_1$ regularization terms:
\begin{equation}
\mathcal{L}_{\text{Total}} = \mathcal{L} + \gamma_s \norm{\vec{w}_s}_1 + \sum_{k=1}^{K} \gamma_k \norm{\vec{w}_k}_1,
\end{equation}
for additional positive hyperparameters $\gamma_s$ and $\gamma_k, k=1,\dots,K$.

\section{Experiments}
We evaluate all new methods against DeepHit \citep{Lee2018DeepHitAD} as well as the most common survival analysis models for competing risks, Fine-Gray models \citep{Fine:1999} and Random Survival Forests (RSF) \citep{Ishwaran:2014}, on two medical datasets described below (details in Appendix \ref{app:datadescription}).

The Prostate, Lung, Colorectal and Ovarian Cancer (\textbf{PLCO}) Screening Trial includes baseline information and prostate cancer screening data (105 features) from 38,052 patients, for which we predict prostate cancer incidence (Event 1), with death prior to prostate cancer incidence (Event 2) a competing risk. The trial has been previously described in \citep{prorok_PLCOdesign_2000}.

We also use an extracted cohort of 72,809 breast cancer patients from the Surveillance, Epidemiology, and End Results (\textbf{SEER}) Program. The patients have baseline medical information (23 features) to predict death from breast cancer (Event 1). Death from cardiovascular disease (Event 2) and death from other causes (Event 3) are treated as competing risks. In order to compare performance when there are a large number of irrelevant features, we also artificially extended the SEER dataset with 20 to 100 synthetic binary features to form the \textbf{SEERsynth} dataset. The additional features are constructed for each patient independently from both other features and the survival distribution. 

We use an extension of the C-index \citep{Gerds:2013aa} to evaluate discriminative performance at various time horizons, and adapt it to the competing risks setting. Given estimates $\hat F_k$ of cumulative incidence functions, patient features $\vec{x}$ and survival times $\tau$, the C-index for event $k$ is
\begin{equation}
C_k(\Delta t) = \probX{ \hat F_k (\Delta t \mid \vec{x}^i) > \hat F_k (\Delta t \mid \vec{x}^j) \bigg \vert \tau^i < \tau^j, k^i = k, \tau^i < \Delta t}.
\end{equation}

We evaluate performance at time horizons $\Delta t$ of 12, 36, 60 and 120 months. The results are presented as averages over 5 train/test splits from 5-fold cross-validation. For DeepHit-based models we run 50 iterations of random search for hyperparameters on the first cross-validation (except for DeepHit, where 20 iterations suffice due to the smaller search space). The optimal parameters are then applied to all other train/test splits (see Appendix \ref{app:implementation} for details).

\begin{table}[h!]
\caption{C-index performance of all models on PLCO and SEERsynth. Bold figures indicate the best performing method for each event and evaluation horizon.}
\label{tbl:small_results}
\centering
\scriptsize
\begin{tabular}{ll|rr|rrr}
\toprule
                 &           & \multicolumn{2}{c}{PLCO} & \multicolumn{3}{c}{SEER (100 synth. features)} \\
 \textbf{Horizon}                & \textbf{Algorithm}          &         Event 1 &         Event 2 &                    Event 1 &         Event 2 &         Event 3 \\
\midrule
\multirow{8}{*}{\textbf{$\Delta t = 012$}} & \textbf{Sparse\DeepHitPlus{}} &           0.930 &  \textbf{0.760} &                      0.846 &           0.698 &           0.763 \\
                 & \textbf{Filter\DeepHitPlus{} (Anova)} &  \textbf{0.935} &           0.753 &                      0.840 &           0.662 &           0.760 \\
                 & \textbf{Filter\DeepHitPlus{} (SVM)} &           0.933 &           0.741 &                      0.809 &           0.696 &           0.748 \\
                 & \textbf{Filter\DeepHitPlus{} (ReliefF)} &           0.931 &           0.747 &                      0.836 &  \textbf{0.709} &  \textbf{0.769} \\
                 & \textbf{\DeepHitPlus{}} &           0.927 &  \textbf{0.760} &                      0.822 &           0.687 &           0.750 \\
                 & \textbf{DeepHit} &           0.906 &           0.746 &                      0.798 &           0.660 &           0.736 \\
                 & \textbf{RSF} &           0.934 &           0.716 &             \textbf{0.852} &           0.616 &           0.747 \\
                 & \textbf{Fine-Gray} &           0.710 &  \textbf{0.760} &                      0.793 &           0.642 &           0.701 \\
\cline{1-7}
\multirow{8}{*}{\textbf{$\Delta t = 060$}} & \textbf{Sparse\DeepHitPlus{}} &           0.869 &  \textbf{0.757} &                      0.758 &           0.678 &  \textbf{0.697} \\
                 & \textbf{Filter\DeepHitPlus{} (Anova)} &  \textbf{0.871} &           0.751 &                      0.754 &           0.651 &  \textbf{0.697} \\
                 & \textbf{Filter\DeepHitPlus{} (SVM)} &           0.870 &           0.744 &                      0.741 &           0.695 &           0.685 \\
                 & \textbf{Filter\DeepHitPlus{} (ReliefF)} &           0.866 &           0.746 &                      0.749 &  \textbf{0.698} &           0.696 \\
                 & \textbf{\DeepHitPlus{}} &           0.845 &           0.756 &                      0.717 &           0.665 &           0.682 \\
                 & \textbf{DeepHit} &           0.807 &           0.744 &                      0.695 &           0.648 &           0.682 \\
                 & \textbf{RSF} &           0.866 &           0.748 &             \textbf{0.770} &           0.680 &           0.693 \\
                 & \textbf{Fine-Gray} &           0.647 &           0.755 &                      0.667 &           0.662 &           0.672 \\
\bottomrule
\end{tabular}

\end{table}

Table \ref{tbl:small_results} shows the C-index results of all methods for the first and third evaluation horizons (see Appendix \ref{app:fullresults} for full results). We find that DeepHit significantly underperforms the traditional algorithm RSF on both datasets. \DeepHitPlus{} delivers improvements through more adaptive layer sizing. Sparse\DeepHitPlus{} achieves a 1.2\% improvement of average C-index over \DeepHitPlus{} on PLCO, and 3.1\% on SEERsynth. Filter\DeepHitPlus{} methods outperform \DeepHitPlus{} on PLCO by 0.3-0.9\%, and 1.6-3.3\% on SEERsynth.

\begin{figure}[h!]
\centering
\includegraphics[width=\textwidth]{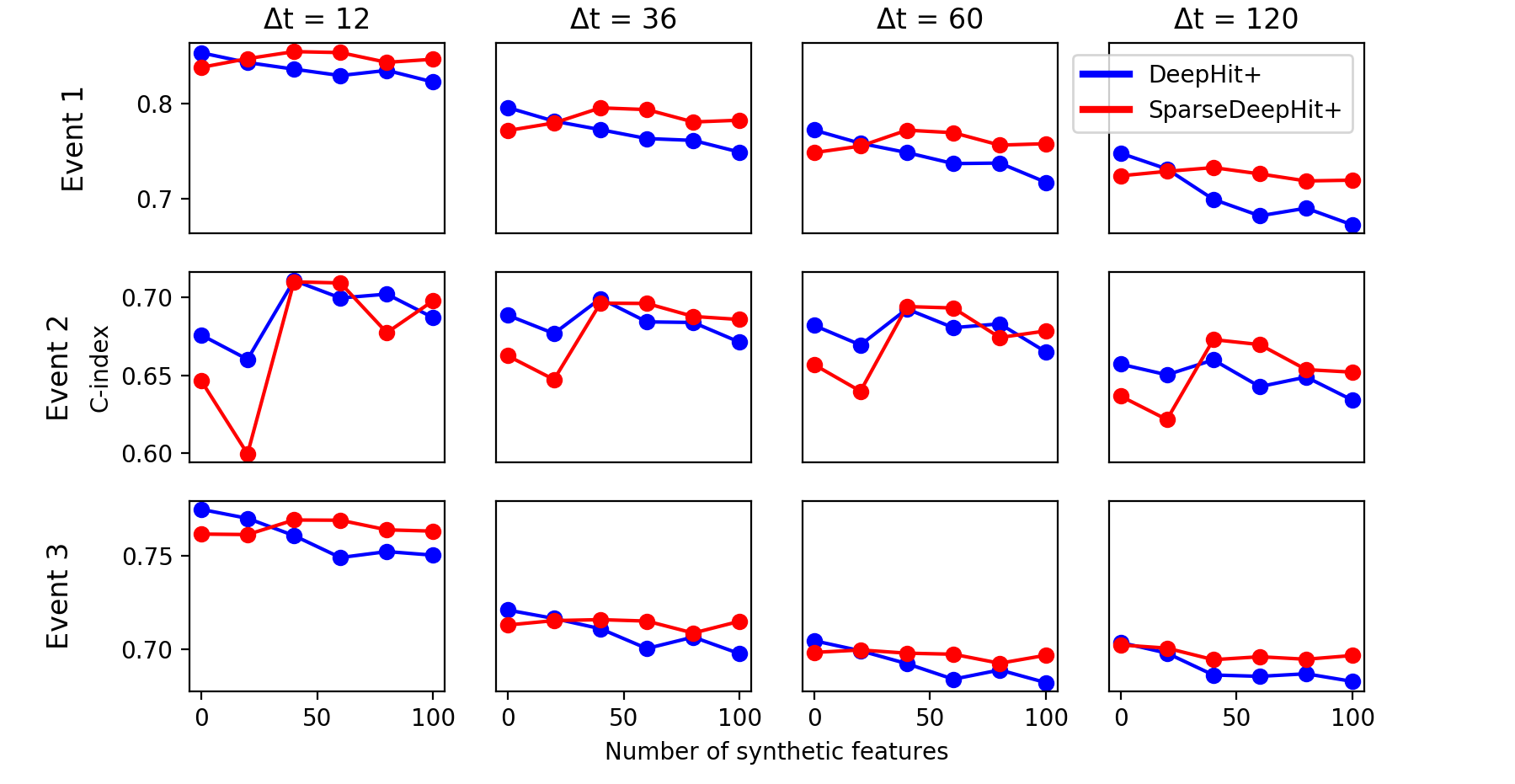}
\caption{\DeepHitPlus{} and Sparse\DeepHitPlus{} on SEER with varying number of synthetic features}
\label{fig:nips_synth_evolution}
\end{figure}

Figure \ref{fig:nips_synth_evolution} shows how the performance of \DeepHitPlus{} decreases when synthetic features are added. This gives evidence that the performance of deep learning survival analysis methods can significantly suffer in settings where the number of potentially irrelevant features becomes large, motivating the need for feature selection applications. In Figure \ref{fig:nips_synth_evolution} we also observe that the performance with feature selection, given by Sparse\DeepHitPlus{}, remains more stable as synthetic features are added. We note that for Event 2 performance is variable due to the small number of observations (1\% of the entire dataset), which results in small samples particularly for validation and testing (see Appendix \ref{app:datadescription} for dataset details).

Another benefit of feature selection is the output of feature rankings that contribute to the model's interpretability. We present these results in Appendix \ref{app:feature_rankings}, where it is shown that on PLCO, the medically relevant PSA level and DRE result are ranked top, whilst on SEERsynth irrelevant synthetic features are mostly correctly filtered out.

\bibliography{bibliography/refs_clean_carl}
\bibliographystyle{plainnat}

\clearpage
\appendix

\section*{Appendix}

\section{Datasets} \label{app:datadescription}

\subsection{PLCO}
The Prostate, Lung, Colorectal and Ovarian (PLCO) Cancer Screening Trial provides a comprehensive dataset to analyse its prostate cancer screening, incidence and mortality results. Its design has been previously described by \cite{prorok_PLCOdesign_2000}. Between 1993 and 2001, participants aged between 55 and 74 were enrolled in 10 US study centers and randomised into a screening and control group. Those in the screening group were offered annual PSA testing for 6 years and annual digital rectal examination for 4 years, whilst those in the control group received normal care. For both groups, all diagnosed cancers and deaths were collected and monitored, whilst screening results were recorded only for the screening group. In this analysis we restrict our dataset to the screening group in order to target predictions based on screening data.

The event of interest is prostate cancer incidence (Event 1). Furthermore, the participant could have been right-censored prior to prostate cancer diagnosis, or have died, which is included as a separate event (Event 2). The number of participants for each event are as follows:
\begin{itemize}
\item \textbf{Total patients:} 38,052
	\begin{itemize}
	\item \textbf{Event 1:} Prostate cancer incidence: 4,425 (12\%)
	\item \textbf{Event 2:} Death prior to prostate cancer incidence: 6,032 (16\%)
	\item \textbf{Censored:} 27,595 (73\%)
	\end{itemize}
\end{itemize}

There are 66 original features in the dataset. After one-hot encoding of categorical variables we obtained 105 feature columns used in the models:
\begin{itemize}
\item Trial entry / year 0 PSA level (1 feature) and DRE result (1 feature - 6 features after one-hot encoding)
\item PLCO background questionnaire (64 features - 99 features after one-hot encoding) (demographics, smoking, (prostate) cancer family history, body type, NSAIDS, diseases, male specifics, prostate surgery)
\end{itemize}

The dataset had 13\% missing values (prior to one-hot encoding), however all but 15 features had missingness of 5\% or below. PSA level at year 0 had missingness of 10\%. The remaining features with high missingness were mostly questions that were left out for large parts of the cohort, such as details around smoking and medical conditions if the patient had indicated that these conditions apply to him.

Missing values were imputed using the mode for binary/categorical features, and the mean for numerical features. All features were furthermore normalised to mean 0 and variance 1 prior to training the models.

For survival analysis, we set time 0 to be the time of the first PSA or DRE screen, whichever occurred later. The average exit months for the trial were 122 months.

\subsection{SEER}
The Surveillance, Epidemiology, and End Results Program (SEER)\footnote{https://seer.cancer.gov/causespecific/} is a publicly available dataset. We use an extracted cohort of 72,809 breast cancer patients from 1992 to 2007. The event of interest is death, which occured either from breast cancer, cardiovascular disease or other causes. Patients living at the end of the study were right-censored. The number of patients by event are as follows:
\begin{itemize}
\item \textbf{Total patients:} 72,809
	\begin{itemize}
	\item \textbf{Event 1:} Death from breast cancer: 10,634 (15\%)
	\item \textbf{Event 2:} Death from cardiovascular disease (CVD): 903 (1\%)
	\item \textbf{Event 3:} Death from other causes: 4,484 (6\%)
	\item \textbf{Censored:} 56,788 (78\%)
	\end{itemize}
\end{itemize}

\paragraph{SEER: Original features}
The SEER dataset contains 23 features:
\begin{itemize}
\item Demographics, including age, race, gender (8 features)
\item Morphology information (lymphoma subtype, histological type, etc.), diagnostic information, therapy information, tumor size and tumor type (15 features)
\end{itemize}

Missing values were imputed using the mean value for real-valued features and the mode for categorical features. All features were furthermore normalised to mean 0 and variance 1 prior to training the models.

\paragraph{SEERsynth: Additional synthetic features}
In order to compare the performance of methods when there are a large number of irrelevant features, we also artificially extended the SEER dataset with up to 100 synthetic features. The additional features are constructed for each patient independently from both other features and the survival distribution (time and event). For each feature $j$ we initially pick a parameter $p_j$, where
\begin{equation}
p_j \sim \text{Unif}[0,1], j=1, \dots, 100.
\end{equation}
Finally, feature values for each patient are drawn from the distribution $\text{Bernoulli}(p_j)$. They therefore represent additional binary random noise irrelevant to the prediction problem. We choose binary features due to their common occurence in medical datasets either as indicators of medical conditions, or one-hot encodings of categorical features.

The SEERsynth dataset thus includes up to 123 features:
\begin{itemize}
\item Original SEER features (23 features)
\item Synthetic features (up to 100 features)
\end{itemize}

The average exit months for the SEER  and SEERsynth datasets were 127 months.

\section{Hybrid\DeepHitPlus{}} \label{app:hybriddeephitplus}
We develop a hybrid feature selection approach that uses the machine learning model itself to extract feature relevance values from the data as an alternative to filter methods. The procedure, using an adaptation of Breiman and Cutler's permutation importance \citep{Breiman2001}, works as follows:
\begin{enumerate}
\item We initially train \DeepHitPlus{} using all features.
\item For each feature we then randomly permute its validation set values, and run the pre-trained model to predict using the new permuted validation set. The importance of the feature is defined to be the difference in true model C-index performance (using original data) and new model performance (using the new data with the permuted feature column). Feature importances are therefore event-specific, depending on the event chosen for the performance calculation. We compute the importance values for all evaluation times, and report the average. Furthermore we average importance values over $I$ permutations.
\item We finally choose the top $m_k$ features according to their importance results, treating the number of features to be selected ($m_k, k=1,\dots,K$) as additional hyperparameters that are optimised with random search during training. The deep neural network architecture is as for Filter\DeepHitPlus{} in Figure \ref{fig:dh_architecture_modified}. The shared sub-network's input features are the common features that are selected for all events.
\end{enumerate}

\section{Implementation and hyperparameter optimization} \label{app:implementation}
\paragraph{Fine-Gray}
We implemented Fine-Gray models using the \lstinline{crr} function in the R package \lstinline{cmprsk}\footnote{https://cran.r-project.org/web/packages/cmprsk/}, using default values for all function parameters. The model fails when there are singularities in the input data. We therefore remove linearly dependent columns from the dataset prior to training the model.

\paragraph{Random Survival Forest}
Random Survival Forests were implemented using the \lstinline{rfsrc} function in the R package \lstinline{randomForestSRC}\footnote{https://cran.r-project.org/web/packages/randomForestSRC/}. The number of trees was chosen to be 1000, and all other parameters were set to defaults.

\paragraph{DeepHit}
We implemented DeepHit (as well as \DeepHitPlus{} and all its extensions) using Python's \lstinline{tensorflow}\footnote{https://www.tensorflow.org/} package. The fixed model settings and layer sizes used were as described in \citep{Lee2018DeepHitAD}, apart from early stopping, which was conducted using validation C-index performance, as opposed to validation loss.
\edef\tensorflowfootnote{\the\value{footnote}} 

The remaining hyperparameters $\beta$ and $\sigma$ were chosen by random search. This was conducted with 20 search iterations on the first cross-validation train/test split (almost amounting to exhaustive search given the small hyperparameter space). We determined the best hyperparameters based on average C-index performance across events and evaluation times on the validation dataset. During the random search procedure, values for the hyperparameters were chosen from the sets in Table \ref{tbl:hyperparamrschoices_initialv0}. The second to fifth cross-validations used the hyperparameters determined from the first split in order to improve computational efficiency. The size of the validation dataset for random search and early stopping was 20\% of the original training data. 

\begin{table}[h!]
\caption{Random search hyperparameter choices for DeepHit}
\label{tbl:hyperparamrschoices_initialv0}
\small
\centering
\begin{tabular}{lc}
\toprule
\thead{Hyperparameter} & \thead{Set of choices} \\ \hline
$\beta$: Weight of ranking loss & 0.1, 0.3, 1, 3, 10 \\
$\sigma$: Parameter for risk comparator $\eta$ & 0.1, 0.3, 1, 3, 10 \\
\bottomrule
\end{tabular}
\end{table}

\paragraph{\DeepHitPlus{}}
Here we conduct additional hyperparameter random search for the number of layers and hidden nodes in each sub-network. In order to focus on these variables, we fix the hyperparameters $\beta$ and $\sigma$ to their optimal values determined by DeepHit's initial run on the same dataset. Given the larger hyperparameter space, we conduct 50 random search iterations on the first cross-validation, and apply the optimal parameters to all other train/test splits. The sets from which the hyperparameters were chosen are given in Table \ref{tbl:hyperparamrschoices_deephitplus}.

\begin{table}[h!]
\caption{Random search hyperparameter choices for \DeepHitPlus{}}
\label{tbl:hyperparamrschoices_deephitplus}
\small
\centering
\begin{tabular}{lc}
\toprule
\thead{Hyperparameter} & \thead{Set of choices} \\ \hline
$n_s$, number of shared layers: & 1, 2, 3 \\
$h_s$, nodes per hidden shared layer:& 50, 100, 200 \\
$n_k$, number of cause-specific layers: & 1, 2, 3 \\
$h_k$, nodes per hidden cause-specific layer:& 50, 100, 200 \\
\bottomrule
\end{tabular}
\end{table}

\paragraph{Filter-, Sparse- and Hybrid\DeepHitPlus{}}
These models are trained in the same way as \DeepHitPlus{}. However, as each extension adds additional parameters, we restrict the choices for network sizing compared to \DeepHitPlus{} in an effort to keep the search space manageable within the computational budget. The sets from which these hyperparameters were chosen are given in Table \ref{tbl:hyperparamrschoices_shared}. In order to reduce the search space for Sparse\DeepHitPlus{}, we set the weighting for the shared sub-network's regularisation term to be $\gamma_s = \frac{1}{K} \sum_{k=1}^{K} \gamma_k$.

\begin{table}[h!]
\caption{Random search hyperparameter choices for Filter-, Sparse- and Hybrid\DeepHitPlus{}}
\label{tbl:hyperparamrschoices_shared}
\small
\centering
\begin{tabular}{lc}
\toprule
\thead{Hyperparameter} & \thead{Set of choices} \\ \hline
$n_s$, number of shared layers: & 1, 2 \\
$h_s$, nodes per hidden shared layer:& 50, 100 \\
$n_k$, number of cause-specific layers: & 1, 2 \\
$h_k$, nodes per hidden cause-specific layer:& 50, 100 \\
$m_k$, number of features selected (Filter-, Hybrid\DeepHitPlus{}): & 20, 40, 60 \\
$\gamma_k$, regularisation loss weights (Sparse\DeepHitPlus{}): & 0.00001, 0.0001, 0.001 \\
\bottomrule
\end{tabular}
\end{table}

\section{Full results} \label{app:fullresults}
We present full results for all algorithms and evaluation horizons in Tables \ref{tbl:fullresults_pros} and \ref{tbl:fullresults_synth1}.

\begin{table}[h!]
\caption{C-index performance of all models on PLCO. SD denotes standard deviation of the 5 train-test splits, $\Delta$ denotes C-index difference to \DeepHitPlus{}.}
\label{tbl:fullresults_pros}
\centering
\scriptsize
\begin{tabular}{ll|rrr|rrr}
\toprule
                 & \textbf{Event} & \multicolumn{3}{c}{Event 1} & \multicolumn{3}{c}{Event 2} \\
                 & {} &            Mean &     SD & $\Delta$ &            Mean &     SD & $\Delta$ \\
\textbf{Horizon} & \textbf{Algorithm} &                 &        &          &                 &        &          \\
\midrule
\multirow{9}{*}{\textbf{$\Delta t = 012$}} & \textbf{Hybrid\DeepHitPlus{}} &  \textbf{0.936} &  0.008 &    0.008 &           0.748 &  0.032 &   -0.012 \\
                 & \textbf{Filter\DeepHitPlus{} (Anova)} &           0.935 &  0.009 &    0.007 &           0.753 &  0.028 &   -0.007 \\
                 & \textbf{RSF} &           0.934 &  0.006 &    0.007 &           0.716 &  0.040 &   -0.045 \\
                 & \textbf{Filter\DeepHitPlus{} (SVM)} &           0.933 &  0.007 &    0.005 &           0.741 &  0.050 &   -0.019 \\
                 & \textbf{Filter\DeepHitPlus{} (ReliefF)} &           0.931 &  0.007 &    0.004 &           0.747 &  0.034 &   -0.013 \\
                 & \textbf{Sparse\DeepHitPlus{}} &           0.930 &  0.008 &    0.003 &  \textbf{0.760} &  0.027 &   -0.000 \\
                 & \textbf{\DeepHitPlus{}} &           0.927 &  0.009 &    0.000 &  \textbf{0.760} &  0.030 &    0.000 \\
                 & \textbf{DeepHit} &           0.906 &  0.013 &   -0.021 &           0.746 &  0.038 &   -0.014 \\
                 & \textbf{Fine-Gray} &           0.710 &  0.020 &   -0.218 &  \textbf{0.760} &  0.028 &   -0.001 \\
\cline{1-8}
\multirow{9}{*}{\textbf{$\Delta t = 036$}} & \textbf{Hybrid\DeepHitPlus{}} &  \textbf{0.890} &  0.004 &    0.020 &           0.752 &  0.014 &   -0.006 \\
                 & \textbf{Filter\DeepHitPlus{} (Anova)} &           0.888 &  0.006 &    0.017 &           0.755 &  0.015 &   -0.003 \\
                 & \textbf{Filter\DeepHitPlus{} (SVM)} &           0.887 &  0.006 &    0.016 &           0.745 &  0.015 &   -0.013 \\
                 & \textbf{Sparse\DeepHitPlus{}} &           0.886 &  0.005 &    0.016 &  \textbf{0.758} &  0.011 &    0.000 \\
                 & \textbf{Filter\DeepHitPlus{} (ReliefF)} &           0.885 &  0.005 &    0.014 &           0.749 &  0.017 &   -0.009 \\
                 & \textbf{RSF} &           0.884 &  0.006 &    0.013 &           0.742 &  0.016 &   -0.016 \\
                 & \textbf{\DeepHitPlus{}} &           0.871 &  0.007 &    0.000 &  \textbf{0.758} &  0.011 &    0.000 \\
                 & \textbf{DeepHit} &           0.835 &  0.010 &   -0.036 &           0.746 &  0.012 &   -0.012 \\
                 & \textbf{Fine-Gray} &           0.666 &  0.021 &   -0.205 &           0.757 &  0.011 &   -0.001 \\
\cline{1-8}
\multirow{9}{*}{\textbf{$\Delta t = 060$}} & \textbf{Hybrid\DeepHitPlus{}} &  \textbf{0.872} &  0.006 &    0.027 &           0.751 &  0.015 &   -0.005 \\
                 & \textbf{Filter\DeepHitPlus{} (Anova)} &           0.871 &  0.006 &    0.025 &           0.751 &  0.016 &   -0.005 \\
                 & \textbf{Filter\DeepHitPlus{} (SVM)} &           0.870 &  0.004 &    0.025 &           0.744 &  0.017 &   -0.012 \\
                 & \textbf{Sparse\DeepHitPlus{}} &           0.869 &  0.004 &    0.024 &  \textbf{0.757} &  0.012 &    0.001 \\
                 & \textbf{Filter\DeepHitPlus{} (ReliefF)} &           0.866 &  0.003 &    0.021 &           0.746 &  0.016 &   -0.010 \\
                 & \textbf{RSF} &           0.866 &  0.006 &    0.020 &           0.748 &  0.011 &   -0.009 \\
                 & \textbf{\DeepHitPlus{}} &           0.845 &  0.008 &    0.000 &           0.756 &  0.012 &    0.000 \\
                 & \textbf{DeepHit} &           0.807 &  0.012 &   -0.038 &           0.744 &  0.013 &   -0.013 \\
                 & \textbf{Fine-Gray} &           0.647 &  0.013 &   -0.198 &           0.755 &  0.013 &   -0.002 \\
\cline{1-8}
\multirow{9}{*}{\textbf{$\Delta t = 120$}} & \textbf{Hybrid\DeepHitPlus{}} &  \textbf{0.821} &  0.005 &    0.032 &           0.738 &  0.002 &   -0.006 \\
                 & \textbf{Sparse\DeepHitPlus{}} &           0.820 &  0.006 &    0.031 &           0.745 &  0.005 &    0.001 \\
                 & \textbf{Filter\DeepHitPlus{} (SVM)} &           0.820 &  0.006 &    0.031 &           0.733 &  0.006 &   -0.011 \\
                 & \textbf{Filter\DeepHitPlus{} (Anova)} &           0.819 &  0.007 &    0.030 &           0.739 &  0.003 &   -0.005 \\
                 & \textbf{Filter\DeepHitPlus{} (ReliefF)} &           0.814 &  0.006 &    0.025 &           0.732 &  0.004 &   -0.012 \\
                 & \textbf{RSF} &           0.813 &  0.005 &    0.025 &           0.737 &  0.007 &   -0.007 \\
                 & \textbf{\DeepHitPlus{}} &           0.789 &  0.008 &    0.000 &           0.744 &  0.004 &    0.000 \\
                 & \textbf{DeepHit} &           0.749 &  0.011 &   -0.039 &           0.726 &  0.007 &   -0.017 \\
                 & \textbf{Fine-Gray} &           0.622 &  0.008 &   -0.167 &  \textbf{0.746} &  0.005 &    0.003 \\
\bottomrule
\end{tabular}

\end{table}

\begin{table}[h!]
    \begin{adjustwidth}{-0.25in}{-0.25in}  
\caption{C-index performance of all models on SEERsynth (100 additional synthetic features). SD denotes standard deviation of the 5 train-test splits, $\Delta$ denotes C-index difference to \DeepHitPlus{}.}
\label{tbl:fullresults_synth1}
\centering
\scriptsize
\begin{tabular}{ll|rrr|rrr|rrr}
\toprule
                 & \textbf{Event} & \multicolumn{3}{c}{Event 1} & \multicolumn{3}{c}{Event 2} & \multicolumn{3}{c}{Event 3} \\
                 & {} &            Mean &     SD & $\Delta$ &            Mean &     SD & $\Delta$ &            Mean &     SD & $\Delta$ \\
\textbf{Horizon} & \textbf{Algorithm} &                 &        &          &                 &        &          &                 &        &          \\
\midrule
\multirow{9}{*}{\textbf{$\Delta t = 012$}} & \textbf{RSF} &  \textbf{0.852} &  0.015 &    0.029 &           0.616 &  0.081 &   -0.071 &           0.747 &  0.046 &   -0.003 \\
                 & \textbf{Hybrid\DeepHitPlus{}} &           0.848 &  0.016 &    0.026 &           0.692 &  0.072 &    0.005 &           0.763 &  0.023 &    0.013 \\
                 & \textbf{Sparse\DeepHitPlus{}} &           0.846 &  0.022 &    0.024 &           0.698 &  0.089 &    0.011 &           0.763 &  0.029 &    0.013 \\
                 & \textbf{Filter\DeepHitPlus{} (Anova)} &           0.840 &  0.018 &    0.018 &           0.662 &  0.124 &   -0.024 &           0.760 &  0.026 &    0.010 \\
                 & \textbf{Filter\DeepHitPlus{} (ReliefF)} &           0.836 &  0.012 &    0.014 &  \textbf{0.709} &  0.084 &    0.022 &  \textbf{0.769} &  0.018 &    0.019 \\
                 & \textbf{\DeepHitPlus{}} &           0.822 &  0.017 &    0.000 &           0.687 &  0.057 &    0.000 &           0.750 &  0.019 &    0.000 \\
                 & \textbf{Filter\DeepHitPlus{} (SVM)} &           0.809 &  0.045 &   -0.013 &           0.696 &  0.094 &    0.009 &           0.748 &  0.030 &   -0.003 \\
                 & \textbf{DeepHit} &           0.798 &  0.021 &   -0.025 &           0.660 &  0.048 &   -0.027 &           0.736 &  0.024 &   -0.014 \\
                 & \textbf{Fine-Gray} &           0.793 &  0.020 &   -0.029 &           0.642 &  0.061 &   -0.045 &           0.701 &  0.015 &   -0.050 \\
\cline{1-11}
\multirow{9}{*}{\textbf{$\Delta t = 036$}} & \textbf{RSF} &  \textbf{0.792} &  0.013 &    0.043 &           0.662 &  0.042 &   -0.009 &           0.704 &  0.024 &    0.007 \\
                 & \textbf{Hybrid\DeepHitPlus{}} &           0.786 &  0.012 &    0.037 &           0.690 &  0.036 &    0.019 &           0.714 &  0.022 &    0.016 \\
                 & \textbf{Sparse\DeepHitPlus{}} &           0.782 &  0.024 &    0.034 &           0.686 &  0.024 &    0.015 &  \textbf{0.715} &  0.021 &    0.017 \\
                 & \textbf{Filter\DeepHitPlus{} (Anova)} &           0.777 &  0.013 &    0.029 &           0.647 &  0.049 &   -0.025 &  \textbf{0.715} &  0.021 &    0.017 \\
                 & \textbf{Filter\DeepHitPlus{} (ReliefF)} &           0.773 &  0.026 &    0.024 &  \textbf{0.702} &  0.024 &    0.031 &           0.714 &  0.019 &    0.017 \\
                 & \textbf{Filter\DeepHitPlus{} (SVM)} &           0.761 &  0.020 &    0.013 &           0.695 &  0.022 &    0.024 &           0.699 &  0.029 &    0.002 \\
                 & \textbf{\DeepHitPlus{}} &           0.749 &  0.006 &    0.000 &           0.671 &  0.016 &    0.000 &           0.697 &  0.023 &    0.000 \\
                 & \textbf{DeepHit} &           0.725 &  0.008 &   -0.024 &           0.652 &  0.036 &   -0.019 &           0.697 &  0.029 &   -0.000 \\
                 & \textbf{Fine-Gray} &           0.697 &  0.009 &   -0.052 &           0.654 &  0.021 &   -0.017 &           0.674 &  0.013 &   -0.023 \\
\cline{1-11}
\multirow{9}{*}{\textbf{$\Delta t = 060$}} & \textbf{RSF} &  \textbf{0.770} &  0.006 &    0.053 &           0.680 &  0.041 &    0.015 &           0.693 &  0.016 &    0.011 \\
                 & \textbf{Hybrid\DeepHitPlus{}} &           0.763 &  0.008 &    0.046 &           0.682 &  0.031 &    0.018 &           0.695 &  0.012 &    0.013 \\
                 & \textbf{Sparse\DeepHitPlus{}} &           0.758 &  0.020 &    0.041 &           0.678 &  0.032 &    0.014 &  \textbf{0.697} &  0.013 &    0.015 \\
                 & \textbf{Filter\DeepHitPlus{} (Anova)} &           0.754 &  0.007 &    0.037 &           0.651 &  0.036 &   -0.014 &  \textbf{0.697} &  0.013 &    0.016 \\
                 & \textbf{Filter\DeepHitPlus{} (ReliefF)} &           0.749 &  0.028 &    0.032 &  \textbf{0.698} &  0.022 &    0.033 &           0.696 &  0.014 &    0.014 \\
                 & \textbf{Filter\DeepHitPlus{} (SVM)} &           0.741 &  0.016 &    0.024 &           0.695 &  0.029 &    0.030 &           0.685 &  0.014 &    0.003 \\
                 & \textbf{\DeepHitPlus{}} &           0.717 &  0.007 &    0.000 &           0.665 &  0.008 &    0.000 &           0.682 &  0.014 &    0.000 \\
                 & \textbf{DeepHit} &           0.695 &  0.010 &   -0.022 &           0.648 &  0.033 &   -0.016 &           0.682 &  0.014 &   -0.000 \\
                 & \textbf{Fine-Gray} &           0.667 &  0.007 &   -0.050 &           0.662 &  0.035 &   -0.002 &           0.672 &  0.011 &   -0.010 \\
\cline{1-11}
\multirow{9}{*}{\textbf{$\Delta t = 120$}} & \textbf{RSF} &  \textbf{0.746} &  0.004 &    0.074 &  \textbf{0.670} &  0.031 &    0.036 &  \textbf{0.697} &  0.012 &    0.015 \\
                 & \textbf{Filter\DeepHitPlus{} (Anova)} &           0.729 &  0.003 &    0.057 &           0.635 &  0.028 &    0.001 &           0.693 &  0.010 &    0.010 \\
                 & \textbf{Hybrid\DeepHitPlus{}} &           0.724 &  0.009 &    0.052 &           0.658 &  0.026 &    0.024 &           0.688 &  0.010 &    0.005 \\
                 & \textbf{Sparse\DeepHitPlus{}} &           0.719 &  0.013 &    0.047 &           0.652 &  0.032 &    0.018 &           0.696 &  0.009 &    0.014 \\
                 & \textbf{Filter\DeepHitPlus{} (ReliefF)} &           0.706 &  0.029 &    0.034 &           0.665 &  0.019 &    0.031 &           0.692 &  0.010 &    0.009 \\
                 & \textbf{Filter\DeepHitPlus{} (SVM)} &           0.701 &  0.011 &    0.029 &           0.660 &  0.023 &    0.026 &           0.685 &  0.009 &    0.003 \\
                 & \textbf{\DeepHitPlus{}} &           0.672 &  0.004 &    0.000 &           0.634 &  0.015 &    0.000 &           0.683 &  0.009 &    0.000 \\
                 & \textbf{DeepHit} &           0.661 &  0.010 &   -0.011 &           0.618 &  0.021 &   -0.016 &           0.684 &  0.010 &    0.002 \\
                 & \textbf{Fine-Gray} &           0.637 &  0.005 &   -0.036 &           0.641 &  0.024 &    0.007 &           0.686 &  0.013 &    0.004 \\
\bottomrule
\end{tabular}

    \end{adjustwidth}
\end{table}

\FloatBarrier

\section{Feature rankings} \label{app:feature_rankings}
Table \ref{tbl:filtertopfeats_anova} gives an example of the feature ranking output of Filter\DeepHitPlus{}. This table, as well as all other feature ranking tables in this section, exhibits results from the first of the five cross-validation train/test splits. For SEERsynth, synthetic features are denoted as \textit{Synth**} (** a placeholder for the feature number). It can be seen that on PLCO, the medically relevant PSA level and DRE result are ranked top, whilst on SEERsynth irrelevant synthetic features are correctly filtered out.

\begin{table}[h!]
\caption{Top 10 features by ANOVA p-value for Event 1 on PLCO and SEERsynth}
\label{tbl:filtertopfeats_anova}
\scriptsize
\centering
\begin{tabular}[t]{lr}
\toprule
PLCO Feature              &    p-value      \\
\midrule
PSALevel             & 7.5e-154 \\
DRE\_AbnormSuspi      &  9.6e-47 \\
DRE\_Negative         &  1.7e-13 \\
Age                  &  3.0e-13 \\
NumRelativesPrCancer &  2.1e-06 \\
FamHistPrCan\_No      &  3.5e-06 \\
FamHistPrCan\_Yes     &  8.0e-06 \\
Occ\_Retired          &  1.4e-04 \\
Occ\_Working          &  3.2e-04 \\
Race\_BlackNonHisp    &  3.7e-04 \\
\bottomrule
\end{tabular}

\hskip 0.5in
\begin{tabular}[t]{lr}
\toprule
SEERsynth Feature                  &      p-value    \\
\midrule
ACJJ Stage               & 1.1e-233 \\
Tumor Marker             &  2.5e-92 \\
EOD 10 Extent            &  2.8e-48 \\
Positive cytology        &  1.3e-33 \\
In situ/Malignant Tumors &  2.2e-29 \\
Histology ICD-O-3        &  2.3e-25 \\
Bilateral                &  1.4e-17 \\
Married                  &  4.4e-17 \\
Single                   &  6.6e-13 \\
Lymphod node             &  7.2e-12 \\
\bottomrule
\end{tabular}

\end{table}

Since the weights of Sparse\DeepHitPlus{}'s first layers are regularized, they also represent an indication of feature relevance. As an example, we present the top 10 for each sub-network for PLCO in Table \ref{tbl:vimp_sparse_pros}, denoting input features as \textit{Shared**} (** a placeholder for the node number) when a cause-specific sub-network utilizes a feature from the shared connection $f_s(\vec{x})$. It can be seen that whilst PSA level and age at trial entry are dominating, further analysis of the weights is more difficult as shared features are highly ranked, but not immediately attributable to original features from the dataset.

\begin{table}[h!]
\caption{Top 10 features by first layer weights in Sparse\DeepHitPlus{}'s sub-networks on PLCO. The ranking is determined by the weight's absolute value.}
\label{tbl:vimp_sparse_pros}
\scriptsize
\centering
\begin{tabular}[t]{lr}
\toprule
{} & Shared \\
Feature            &        \\
\midrule
PSALevel           &  0.551 \\
EnlargedProsBPH    &  0.141 \\
Marital\_Married    &  0.137 \\
AgeInflamedPros    & -0.127 \\
UrinateNight\_Never &  0.122 \\
Prostatectomy      & -0.119 \\
Marital\_NeverMarr  & -0.117 \\
PersCanHist\_Unkn   & -0.116 \\
Hypertension       &  0.114 \\
GallbladderStones  & -0.111 \\
\bottomrule
\end{tabular}

\hfill
\begin{tabular}[t]{lr}
\toprule
{} & Event 1 \\
Feature      &         \\
\midrule
PSALevel     &  -0.577 \\
Age          &   0.243 \\
Shared20     &   0.201 \\
Shared49     &   0.198 \\
Shared42     &  -0.192 \\
Shared38     &   0.191 \\
Shared25     &   0.185 \\
Cig\_Current  &  -0.182 \\
Hypertension &  -0.173 \\
HeartAttack  &   0.173 \\
\bottomrule
\end{tabular}

\hfill
\begin{tabular}[t]{lr}
\toprule
{} & Event 2 \\
Feature           &         \\
\midrule
PSALevel          &   0.765 \\
Age               &   0.342 \\
DRE\_AbnormSuspi   &   0.272 \\
PersCanHist\_No    &  -0.239 \\
Shared29          &  -0.238 \\
Shared42          &   0.237 \\
DurationSmokedCig &   0.234 \\
Shared35          &   0.213 \\
Occ\_Disabled      &   0.204 \\
SmokingPackYears  &   0.204 \\
\bottomrule
\end{tabular}

\end{table}

Hybrid\DeepHitPlus{}'s feature importance values are by construction a direct representation of their relevance for the model's prediction. As can be seen in Table \ref{tbl:vimp_m_deephit}, the top features overlap with those chosen by the ANOVA filter feature selection from Table \ref{tbl:filtertopfeats_anova}. 

\begin{table}[h!]
\caption{Top 10 features by permutation importance for Event 1 on PLCO and SEERsynth}
\label{tbl:vimp_m_deephit}
\scriptsize
\centering
\begin{tabular}[t]{lrr}
\toprule
PLCO Feature           &   Importance     \\
\midrule
PSALevel          & 0.2692 \\
DRE\_AbnormSuspi   & 0.0085 \\
Cig\_Former        & 0.0071 \\
DRE\_NotDoneExp    & 0.0037 \\
DurationSmokedCig & 0.0037 \\
Race\_Asian        & 0.0029 \\
HadProstateBiopsy & 0.0028 \\
DRE\_Negative      & 0.0026 \\
Race\_BlackNonHisp & 0.0022 \\
ProsSurgeries\_No  & 0.0020 \\
\bottomrule
\end{tabular}

\hskip 0.5in
\begin{tabular}[t]{lrr}
\toprule
SEERsynth Feature                  &     Importance   \\
\midrule
ACJJ Stage               & 0.0794 \\
Lymphod node             & 0.0723 \\
Tumor Marker             & 0.0312 \\
Single                   & 0.0043 \\
Histology ICD-O-3        & 0.0029 \\
In situ/Malignant Tumors & 0.0019 \\
Synth22                  & 0.0018 \\
Positive cytology        & 0.0015 \\
Synth96                  & 0.0015 \\
Positive histology       & 0.0015 \\
\bottomrule
\end{tabular}

\end{table}

\end{document}